\documentclass[letterpaper]{article} 
\usepackage{aaai2026}  
\usepackage{times}  
\usepackage{helvet}  
\usepackage{courier}  
\usepackage[hyphens]{url}  
\usepackage{graphicx} 
\usepackage{natbib}  
\usepackage{caption} 
\usepackage{algorithm}
\usepackage{algorithmic}
\usepackage{amsmath}
\usepackage{booktabs}
\usepackage{multirow}
\usepackage{amsfonts}
\usepackage{xcolor}
\usepackage{newfloat}
\usepackage{listings}
\DeclareCaptionStyle{ruled}{labelfont=normalfont,labelsep=colon,strut=off} 
\floatstyle{ruled}
\newfloat{listing}{tb}{lst}{}
\floatname{listing}{Listing}
\title{AirDDE: Multifactor Neural Delay Differential Equations for Air Quality Forecasting}
\author{
    Binqing Wu \textsuperscript{\rm 1,2,\equalcontrib}, Zongjiang Shang \textsuperscript{\rm 1,2,\equalcontrib}, Shiyu Liu \textsuperscript{\rm 3}, Jianlong Huang \textsuperscript{\rm 1,2}, Jiahui Xu \textsuperscript{\rm 1,2}, \\ Ling Chen \textsuperscript{\rm 1,2,}\thanks{Corresponding author.}
}
\affiliations{
    \textsuperscript{\rm 1} State Key Laboratory of Blockchain and Data Security, Zhejiang University\\
    \textsuperscript{\rm 2} College of Computer Science and Technology, Zhejiang University\\
    \textsuperscript{\rm 3} State Key Laboratory of Clean Energy Utilization, Zhejiang University\\
    \{binqingwu, zongjiangshang, 22251226, xujiahui19, lingchen\}@cs.zju.edu.cn, shiyuliu@zju.edu.cn
}

\begin{document}

\maketitle

\begin{abstract}
Accurate air quality forecasting is essential for public health and environmental sustainability, but remains challenging due to the complex pollutant dynamics.
Existing deep learning methods often model pollutant dynamics as an instantaneous process, overlooking the intrinsic delays in pollutant propagation. 
Thus, we propose AirDDE, the first neural delay differential equation framework in this task that integrates delay modeling into a continuous-time pollutant evolution under physical guidance.
Specifically, two novel components are introduced: (1) a memory-augmented attention module that retrieves globally and locally historical features, which can adaptively capture delay effects modulated by multifactor data; and (2) a physics-guided delay evolving function, grounded in the diffusion-advection equation, that models diffusion, delayed advection, and source/sink terms, which can capture delay-aware pollutant accumulation patterns with physical plausibility. 
Extensive experiments on three real-world datasets demonstrate that AirDDE achieves the state-of-the-art forecasting performance with an average MAE reduction of 8.79\% over the best baselines. The code is available at https://github.com/w2obin/airdde-aaai.
\end{abstract}


\section{Introduction}
Rapid industrialization and urbanization over the past decades have exacerbated air pollution, making air quality a critical concern for public health and environmental sustainability \cite{azimi2024unveiling,geng2025fuxi}. 
This pressing issue underscores the importance of accurate air quality forecasting.
Nevertheless, such forecasting is highly challenging owing to the complex pollutant dynamics \cite{vallero2025fundamentals,bodnar2025foundation}.

Many methods have been proposed to tackle this problem in the past decades.
Traditional methods, e.g., physical-chemical simulation models \cite{xie2005impact} and shallow machine learning techniques \cite{lee2012seasonal}, often rely on simplified assumptions and handcrafted features, limiting their ability to capture the latent pollutant dynamics inherent in air quality data.
This limitation has driven a paradigm shift toward deep learning methods, which offer stronger representational capacity \cite{qi2018deep,wu2024weathergnn}.
Early deep learning methods primarily utilize convolutional neural networks (CNNs) \cite{yan2021multi,chen2023deep} and recurrent neural networks (RNNs) \cite{xu2019multitask,xu2021highair} to extract spatial and temporal features of air quality. Building on these foundations, recent works have adopted spatial temporal graph neural networks (STGNNs) \cite{chen2023group,han2023kill} and attention mechanisms \cite{liang2023airformer,xia2025dynamic} to more effectively model spatial and temporal dependencies, enabling more comprehensive feature extraction. 
Despite their advancements, these methods formulate pollutant dynamics as a discrete-time process, where pollutant transitions occur only at fixed temporal intervals. This formulation limits their ability to capture the continuous-time dynamics of pollutants in the real world.

More recently, several methods have explored modeling pollutant dynamics as a continuous-time process by integrating multifactor field data \cite{airradar} and leveraging neural ordinary differential equations (NODEs) \cite{hettige2024airphynet, tian2024air}. 
However, these methods often adopt an instantaneous assumption, wherein the system’s evolution depends only on the current state. This simplification neglects the transmission time, i.e., delay, during pollutant propagation.
In fact, delays are pervasive and essential in real-world air quality systems \cite{cai2021spatial,millgnn}. For example, pollutants emitted in one location may take several hours to be transported by wind before affecting air quality in downstream locations, introducing a non-negligible delay between emission and observable impact.

To model delays in a continuous-time process, a natural approach is to employ neural delay differential equations (NDDEs) \cite{zhuneural,long2024unveiling}. NDDEs are an extension of NODEs, which allow the system's evolution to depend not only on the current state but also on historical states.
While theoretically appealing, existing NDDEs are constrained to modeling uniform delays, i.e., applying the same delay across all spatial locations, which limits their ability to capture location-specific delays.

Despite the heterogeneity of delays being particularly essential and evident, it is non-trivial to capture them effectively.
(1) Delays are modulated by multiple factors. The transport path and arrival time of pollutants are strongly influenced by varying meteorological and geographical factors, e.g., pollutant concentration levels, wind fields, and geographical distances.
(2) Delay effects exhibit spatiotemporal accumulation. The pollutant concentration at a given location and time is not only determined by local pollutants but also by pollutants from other locations, each arriving with different delays. These cumulative effects are grounded in atmospheric dynamic processes, which cannot be captured by existing statistics-based or purely data-driven delay modeling techniques \cite{jiang2023pdformer,long2024unveiling}.

To this end, we propose AirDDE, a multifactor neural delay differential equation framework for air quality forecasting. 
To the best of our knowledge, AirDDE is the first physics-guided work that integrates delay modeling into pollutant continuous-time evolution.
The main contributions are summarized as follows:

\begin{itemize}
    \item We introduce a memory-augmented attention (MAA) module. Given the delay-aware transport paths constructed from geographic distances and real-time wind fields, this module adopts a dual attention mechanism to retrieve globally and locally historical features. 
    Such a design enables MAA to capture multifactor-modulated delay effects fully considering spatial heterogeneity.
    \item We introduce a physics-guided delay evolving (PDE) function. 
    Guided by the diffusion-advection equation, this function models diffusion, delayed advection, and source/sink terms from multifactor features to capture continuous-time pollutant evolution.
    Such a design enables PDE to capture delay-aware pollutant accumulation patterns in a physically consistent manner.
    \item We compare AirDDE with 19 competitive baselines on 3 real-world datasets. The results demonstrate that AirDDE achieves the state-of-the-art (SOTA) performance with an average MAE reduction of 8.79\% over the best baselines.
\end{itemize}

\section{Related Work}
\textbf{Deep learning for air quality modeling.}
Early deep learning methods use CNNs and RNNs to extract spatial and temporal features. For example, DAL \cite{qi2018deep} and AirNet \cite{yu2020airnet} combine CNNs with RNNs to forecast and calibrate air quality index (AQI) measurements, respectively. FAIRY \cite{chen2023deep} utilizes SegNets to learn multiresolution spatial features for air quality estimation. 
Recent studies adopt STGNNs \cite{han2022semi,liang2022mixed} and attention mechanisms \cite{liang2023airformer,geng2025fuxi} for richer dependency modeling.
In the STGNN paradigm, for example, PM2.5GNN \cite{wang2020pm2}, MasterGNN and MasterGNN+ \cite{han2021joint,han2023kill}, and GAGNN \cite{chen2023group} enhance spatial and temporal dependency learning through GNN–GRU integration, adversarial training, and hierarchical graph design, respectively. Parallel to the STGNN-based advances, attention-based methods are also a powerful paradigm \cite{qiu2024tfb,qiu2025duet}. For example, AirFormer \cite{liang2023airformer} proposes a dartboard-style spatial attention and a causal temporal attention for long-term forecasting. AirRadar \cite{airradar} introduces a masked feature reconstruction framework using spatial attention and temporal causal adjustment to infer air quality. Fuxi-Air \cite{geng2025fuxi} leverages a Transformer architecture for air pollution forecasting.
Nevertheless, these methods model pollutant dynamics as a discrete-time process, overlooking their continuous-time nature.

To address this limitation, some recent methods model pollutant dynamics as a continuous-time process by integrating multifactor field data and leveraging NODEs. For example, STFNN \cite{feng2024spatio} unifies field- and graph-based views for fine-grained continuous spatiotemporal inference. AirPhyNet \cite{hettige2024airphynet} embeds pollutant transport equations into a GNN–NODE framework, while AirDualODE \cite{tian2024air} uses dual-branch NODEs combining data-driven and physics-informed components. However, these methods often adopt an instantaneous assumption, which largely ignore propagation delays. Although a few methods in general spatiotemporal tasks model delays using shared patterns \cite{jiang2023pdformer} or cross-correlations \cite{long2024unveiling}, they assume globally uniform delays and fail to capture delays shaped by heterogeneous conditions.

Thus, we propose AirDDE, a physics-guided framework that integrates delay modeling into continuous-time pollutant evolution. We introduce a memory-augmented attention and a delay-aware evolving function to model delay effects conditioned on location- and time-specific multifactor data.
These designs enable AirDDE to make accurate forecasts with strong physical plausibility.

\section{Preliminary}
\noindent \textbf{Task Formulation.}
Given historical air quality observations and auxiliary factors (e.g., meteorological and geographical variables), the goal of air quality forecasting is to predict future air quality for the next time steps, formulated as:
\begin{equation}
    \hat{\boldsymbol{X}}^{T+1:T+H} = \mathcal{F}(\boldsymbol{X}^{1:T},\boldsymbol{M}^{1:T}; \boldsymbol{\Theta}),
\end{equation}
where $\boldsymbol{X}^{1:T} \in \mathbb{R}^{N \times T}$ and $\boldsymbol{M}^{1:T} \in \mathbb{R}^{N \times T}$ represent historical air quality and auxiliary factors from $N$ locations over past $T$ steps, respectively. $\hat{\boldsymbol{X}}^{T+1:T+H} \in \mathbb{R}^{N \times H}$ represents denotes the predicted air quality for the next $H$ steps. $\mathcal{F}$ is the neural network. $\boldsymbol{\Theta}$ is the learnable parameters of $\mathcal{F}$.


\noindent \textbf{Diffusion-Advection Equation.}
The diffusion–advection equation describes the transport of a substance (e.g., pollutants) in a fluid \cite{moreira1998analytical}, combining diffusion, advection, and source/sink effects, formulated as:
\begin{equation}
\frac{\partial u}{\partial t}+\vec{v} \cdot \nabla u=D \nabla^2 u + S ,
\end{equation}
where $\frac{\partial u}{\partial t}$ represents the partial derivative of $u$ with respect to time.
$\vec{v}$ is the velocity vector field. $\nabla u$ is the gradient of $u$. $\vec{v} \cdot \nabla u$ represents the advection term of $u$. $D$ is the diffusion coefficient. $\nabla^2 u$ is the Laplacian of $u$. $D \nabla^2 u$ represents the diffusion term. $S$ represents the source/sink term.

Recent studies \cite{hettige2024airphynet,tian2024air} have modified this equation as a diffusion-advection equation on graphs. The Laplacian operators for advection and diffusion are approximated using the Chebyshev GNN \cite{defferrard2016convolutional}. 

\noindent \textbf{Neural Delay Differential Equations.}
NDDEs \cite{zhuneural,long2024unveiling} extend NODEs \cite{chen2018neural,li2025symbolic} by modeling delayed dynamics, where the state evolution depends not only on the current state but also on historical states. The general form of an NDDE is formulated as:
\begin{equation}
    \boldsymbol{F}: \frac{d\boldsymbol{h}^t}{dt} = f(\boldsymbol{h}^t, \boldsymbol{h}^{t - \tau}),
\end{equation}
where $\boldsymbol{h}^t$ is the state at time step $t$, $\tau$ is a delay, and $f(\cdot)$ is a neural network-based evolution function. Compared to NODEs, NDDEs face more complex initial value and integration problems, but are more effective for delay effects. 

NDDEs are trained by solving $\boldsymbol{F}$ forward and optimizing neural network parameters using automatic differentiation, often with the adjoint sensitivity method. Implementation tools, e.g., torchdiffeq \cite{kidger2021hey}, facilitate efficient simulation and backpropagation.

\section{Methodology}
\subsection{Overview}
\begin{figure}
    \centering
    \includegraphics[width=\linewidth]{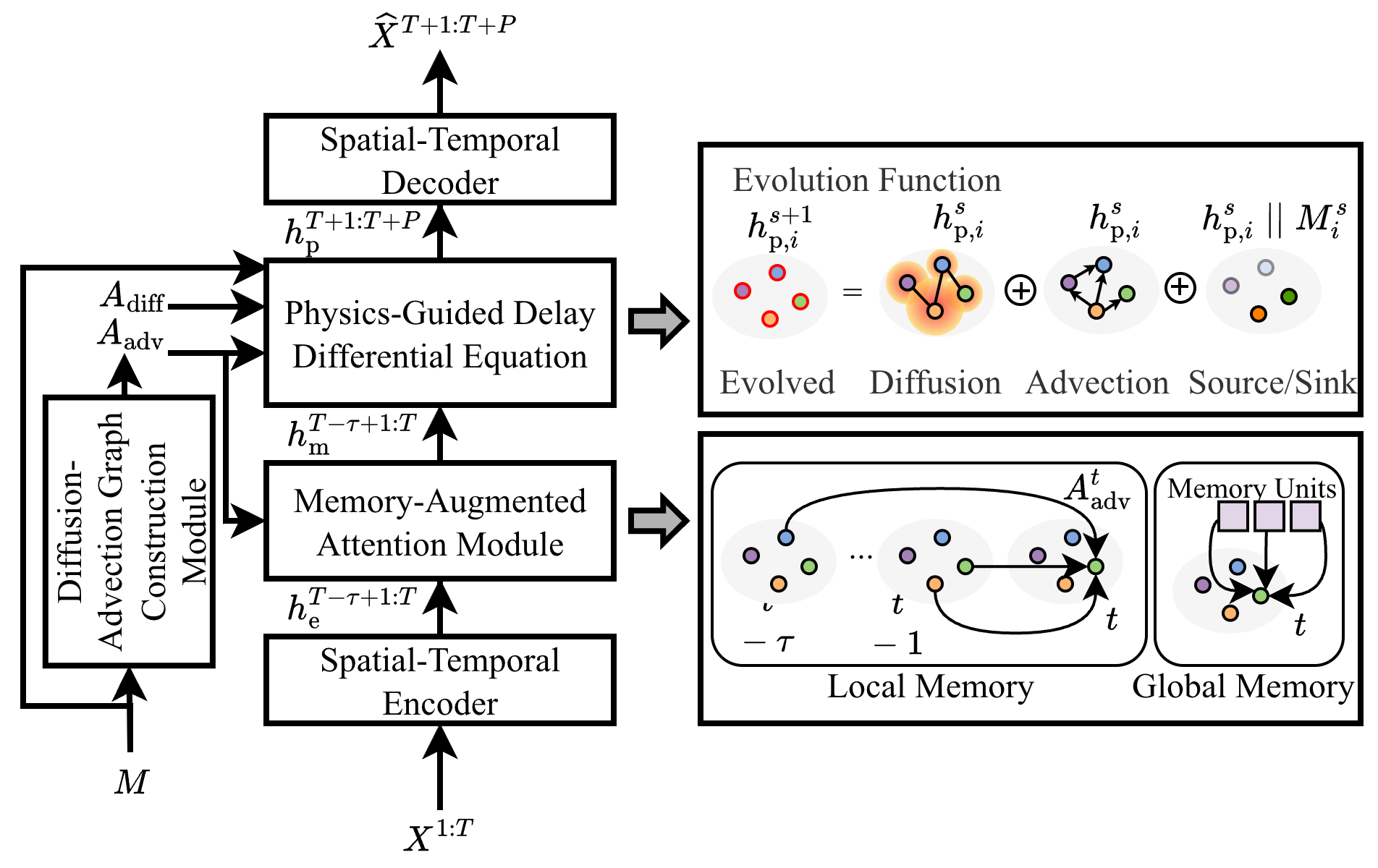}
    \caption{The architecture of AirDDE.}
    \label{fig:framework}
    \vspace{-15px}
\end{figure}

The architecture of AirDDE is illustrated in Fig. \ref{fig:framework}.
AirDDE models continuous-time pollutant propagation with delayed effects via a multifactor-enhanced NDDE framework. 
Considering the complex diffusion and advection effects in air quality, AirDDE constructs diffusion and advection graphs based on wind fields and distances, which can indicate pollutant transport paths. 
Given encoded inputs via an STGNN-based encoder and constructed graphs, AirDDE uses the MAA module to obtain initial states that retrieve global and local historical features, which can adaptively capture delay effects modulated by multifactor data.
Then, AirDDE formulates the PDE function guided by the diffusion-advection equation, which can model delay-aware pollutant accumulation patterns with physical consistency. 
Given the historical states and evolution function, AirDDE adopts a DDE solver that maintains a history buffer to account for delayed states and employs numerical integration to get future air quality states.
After that, these states are fed into a decoder to generate the final predictions.

\subsection{Spatiotemporal Encoder}
To capture spatiotemporal features, we follow an STGNN-based paradigm \cite{jiang2023spatio, chen2024signed}. Specifically, we derive the graph structure from learnable node embeddings and utilize GNNs to replace the MLPs in the GRU's gating mechanisms \cite{bai2020adaptive,wu2023dstcgcn}. This design facilitates adaptive feature extraction by incorporating the underlying graph topology into the temporal updates. The process is formulated as:
\begin{equation} \label{encoder}
\begin{aligned}
\boldsymbol{A} &= \operatorname{SoftMax}(\operatorname{ReLU}(\boldsymbol{E}_{1}(\boldsymbol{E}_{2})^{\text{T}})),\\
\boldsymbol{h}_{\text{e}}^t &= \operatorname{GNN-GRU}(\boldsymbol{X}^t, \boldsymbol{h}_{\text{e}}^{t-1},\boldsymbol{A}) ,
\end{aligned}
\end{equation}
where $\boldsymbol{A} \in \mathbb{R}^{N \times N}$ represents the underlying adjacency matrix, ${\boldsymbol {E}_1}, {\boldsymbol E_2} \in \mathbb{R}^{N \times d} $ are two parameterized node embeddings with $d$ dimensions. $\operatorname{GNN-GRU}(\cdot)$ denotes a GRU variant where the original MLPs in the update and reset gates are replaced with GNNs. $\boldsymbol{X}^t$ and $\boldsymbol{h}_{\text{e}}^{t-1}$ are the current inputs and historical hidden features, respectively.

\subsection{Diffusion-Advection Graph Construction}
\noindent \textbf{Diffusion Graph.} 
Diffusion is significant for pollutant transport, especially when wind is absent or negligible. Since the impact of diffusion is highly related to geographical proximity, we construct a diffusion graph $\boldsymbol{A}_{\text{diff}} \in \mathbb{R}^{N\times N}$ based on Haversine distances between locations, computed from their longitudes and latitudes. The resulting graph is then normalized using a Gaussian kernel \cite{li2017diffusion}.

\noindent \textbf{Advection Graph.}
Advection governs pollutant transport under windy conditions, carrying pollutants across space with inherent delays.
Unlike previous works \cite{jiang2023pdformer,jiang2023spatio}, which rely on globally averaged delays, we construct an advection graph at each time step to capture delays derived from wind fields and geographical distances.
Specifically, a directed edge from location $j$ at time step $t_2$ to location $i$ at time step $t_1$ exists if the wind speed at $j$ at $t_2$ enables an air parcel to reach $i$ at $t_1$. The time lag
$\tau = t_1 - t_2$ represents the travel time of the pollutant from $j$ to $i$, serving as a looking-back window to identify dependencies. The construction is formulated as:
\begin{equation}
\boldsymbol{A}_{\text{adv},ij}^{t}= 
\begin{cases}
1, & \text{if } v_{j}^{\tau} \operatorname{cos}(\theta_{j}^{\tau}) \cdot \tau \geq d_{ij}\\
0, & \text{otherwise}
\end{cases}
\end{equation}
where $v_{j}^{\tau}$ and $\theta_{j}^{\tau}$ represent the wind speed and wind direction at location $j$ at the previous $\tau$ time step, respectively. $d_{ij}$ represents the distance between location $i$ and $j$. $\boldsymbol{A}_{\text{adv}}^{t} \in \mathbb{R}^{N \times N}$ represents the delay-aware dependencies at time step $t$. Notably, $\boldsymbol{A}_{\text{adv}}^{t}$ offers greater adaptiveness than unified or implicit delay assumptions, as it explicitly models delays conditioned on location- and time-specific dynamics.

\subsection{Memory-Augmented Attention Module}
The MAA module is introduced to capture delay-aware initial states modulated by multifactor data.
Pollution propagation shows dual-scale historical patterns: global background trends (e.g., persistent high-PM2.5 regions) and local transient events (e.g., sudden AQI spikes from dust storms). To model these dual-scale delay effects under dynamic multifactor conditions, MAA employs a dual attention mechanism that retrieves both global and local historical features.

\noindent \textbf{Global Memory Modeling.} To capture global historical patterns, we introduce a set of learnable global memory units denoted as $\boldsymbol{M}_{\text{g}} \in \mathbb{R}^{m \times d_{\text{e}}}$, where $m$ is the number of memory units and $d_{\text{e}}$ is the hidden dimension. These memory units aim to memorize global historical patterns, which are randomly initialized and updated during training. We integrate the patterns to the current hidden features $\boldsymbol{h}_{\text{e}}^t$ via attention \cite{vaswani2017attention}, which is formulated as: 
\begin{equation}
    \boldsymbol{h}^t_{\text{g}} = \operatorname{Attention}(\boldsymbol{h}_{\text{e}}^t, \boldsymbol{M}_{\text{g}}, \boldsymbol{M}_{\text{g}}),
\end{equation}
where $\boldsymbol{h}_{\text{e}}^t$ and $\boldsymbol{M}_{\text{g}}$ are linearly projected to obtain the query, keys, and values, respectively.
$\operatorname{Attention}(\cdot)$ first computes similarity scores between query and keys to produce attention weights, which are then used to aggregate values.
This allows $\boldsymbol{h}_{\text{e}}^t$ to adaptively incorporate relevant global memories from $\boldsymbol{M}_{\text{g}}$.


\noindent \textbf{Local Memory Modeling.} To capture local historical patterns, we leverage advection graphs to define dynamic neighborhoods based on wind-driven pollutant transport. Specifically, for location $i$ at time step $t$, we define its neighbors $\mathcal{N}(i)^t$ based on the advection graph $\boldsymbol{A}_{\text{adv}}^{t}$. We then attend over its own and its neighbors’ historical features within a time lag $\tau$, which is formulated as:
\begin{equation}
    \boldsymbol{h}^t_{\text{l},i} = \operatorname{MLP}(\operatorname{Attention}(\boldsymbol{h}_{\text{e},i}^{t},\boldsymbol{h}^{t-\tau+1:t}_{\text{e},j\in{\mathcal{N}(i)^t}},\boldsymbol{h}^{t-\tau+1:t}_{\text{e},j\in{\mathcal{N}(i)^t}})) ,
\end{equation}
where $\boldsymbol{h}_{\text{e},i}^t$ is used as the query, while $\boldsymbol{h}^{t-\tau+1:t}_{\text{e},j\in{\mathcal{N}(i)^t}} \in \mathbb{R}^{|\mathcal{N}(i)^t| \times \tau \times d}$ serves as keys and values. $\boldsymbol{h}^t_{\text{l},i} \in \mathbb{R}^{d}$ is the output after an MLP layer.

We then concatenate the hidden features with those integrating global and local historical patterns and embed them for the initial states. The process is formulated as $\boldsymbol{h}_{\text{m}}^t = \operatorname{MLP}(\operatorname{concat}(\boldsymbol{h}_{\text{e}}^t,\boldsymbol{h}_{\text{g}}^t,\boldsymbol{h}_{\text{l}}^t))$, where $\boldsymbol{h}_{\text{m}}^t \in \mathbb{R}^{N \times d}$ is the output features of the MAA module, which comprehensively capture dual-scale delay effects considering multifactor data.

\subsection{Physics-Guided Delay Evolving Function}
The PDE function is introduced to formulate the delay-aware evolution with physical consistency.
Unlike prior works that assume conservative pollutant transport, we capture real-world non-conservative behavior, e.g., windborne inflows for sources and precipitation-driven removal for sinks. We model such hidden source/sink dynamics from multifactor features, as different factors encode complementary physical signals that together reveal the latent drivers of pollutant variation. Formally, the evolution function for the pollutant is formulated as:
\begin{equation}
\begin{aligned}
\boldsymbol{F}:\frac{d\boldsymbol{h}^t}{dt} & = 
D \cdot \operatorname{GNN}_{\text{diff}}(\boldsymbol{A}_{\text{diff}},\boldsymbol{h}^t)  \\
& + \operatorname{GNN}_{\text{adv}}(\boldsymbol{A}_{\text{adv}}^{t}, \boldsymbol{h}^{t-\tau})
+ f(\boldsymbol{h}^t||\boldsymbol{M}) ,
\end{aligned}
\end{equation}
where $\operatorname{GNN}(\cdot)$ denotes the $K$-hop message passing mechanism that approximates $K$-order Chebyshev GNN, which can achieve a better efficiency and adaptability for dynamics \cite{hamilton2020graph}. $D$ is the diffusion coefficient, which is empirically set to $0.1$ \cite{hettige2024airphynet}. $f$ is a lightweight MLP-based network. $\boldsymbol{M}$ is the multifactor features.

Given the delay-aware initial states $\boldsymbol{h}_{\text{m}}^T$ and historical states $\boldsymbol{h}_{\text{m}}^{T-\tau:T-1}$, the hidden states of pollutant concentrations from $T+1$ to $T+P$ can be obtained by solving $\boldsymbol{F}$. The process is formulated as:
\begin{equation} \label{solver}
\boldsymbol{h}^{T+1: T+P}_{\text{p}} =\text {DDESolver}(\boldsymbol{F}, \boldsymbol{h}^T_{\text{m}}, \boldsymbol{h}^{T-\tau: T}_{\text{m}}, \boldsymbol{M}) ,
\end{equation}
where the solver, following the existing works \cite{chen2018neural,long2024unveiling}, is implemented by torchdiffeq \cite{kidger2021hey}. The solver maintains an explicit memory set of past states during Fourth-order Runge-Kutta integration. 
By incorporating current and past states grounded in the diffusion-advection equation, the future states offer a more physically consistent representation of delay-aware air quality evolution.

\begin{table}[]
    \centering
    \resizebox{\linewidth}{!}{
    \begin{tabular}{c|cccc}
    \toprule
        Dataset & \# Factors & \# Locations &  Time Range & Granularity \\
    \midrule
        KnowAir & 18 & 184 &  1.1.2015-12.31.2018 & 3h \\
        China-AQI & 8 & 209 & 1.1.2017-4.30.2019 & 1h \\
        US-PM & 8 & 175 & 1.1.2020-12.31.2021 & 1h \\
    \bottomrule
    \end{tabular}}
    \caption{Dataset statistics.}
    \label{tab:dataset}
    \vspace{-15px}
\end{table}

\begin{table*}[]
\centering
\resizebox{0.7 \textwidth}{!}{
\begin{tabular}{c|c|ccc|ccc|ccc}
\toprule
\multicolumn{2}{c|}{Dataset} & \multicolumn{3}{c|}{KnowAir} & \multicolumn{3}{c|}{China-AQI} & \multicolumn{3}{c}{US-PM} \\ \midrule
\multicolumn{2}{c|}{Metric} & MAE & RMSE & SMAPE & MAE & RMSE & MAPE & MAE & RMSE & MAPE \\
\midrule
\multirow{8}{*}{\rotatebox{90}{STGNNs}} & DCRNN (2018) & 24.02$^*$ & 37.87$^*$ & 0.53$^*$ & 24.78 & 38.05 & 35.76 & 7.49 & 9.72 & 15.85 \\
 & STGCN (2018) & 23.64$^*$ & 32.48$^*$ & 0.52$^*$ & 23.87 & 37.29 & 35.03 & 6.99 & 8.81 & 14.72 \\
 & ASTGCN (2019) & 19.92$^*$ & 31.39$^*$ & 0.44$^*$ & 21.91 & 36.02 & 34.28 & 4.86 & 7.53 & 12.93 \\
 & MTGNN (2020) & 18.92 & 30.34 & 0.41 & 21.56 & 35.80 & 34.65 & 4.63 & 7.03 & 12.85 \\
 & PM25GNN (2020) & 19.32$^*$ & 30.12$^*$ & 0.43$^*$ & 22.01 & 36.21 & 34.12 & 5.24 & 7.86 & 13.43 \\
 & GAGNN (2023) & 20.71 & 32.88 & 0.42 & 19.54 & 33.37 & 32.95 & 4.32 & 6.53 & 12.43 \\
 & MegaCRN (2023) & 18.77 & 29.45 & 0.42 & 18.93 & 32.41 & 32.23 & 3.85 & 5.41 & 11.72 \\
 & HimNet (2024) & 20.98 & 33.00 & 0.44 & 20.72 & 33.78 & 34.08 & 5.28 & 7.79 & 13.28 \\ 
 \midrule
\multirow{5}{*}{\rotatebox{90}{Attentions}} & Corrformer (2023) & 21.11 & 32.97 & 0.44 & 21.22 & 34.93 & 34.97 & 5.95 & 7.71 & 14.37 \\
 & AirFormer (2023) & 19.17$^*$ & 30.19$^*$ & 0.43$^*$ & 19.60 & 33.14 & 32.86 & 3.90 & 5.37 & \underline{11.55} \\
 & PDFormer (2023) & 19.06 & 30.66 & \underline{0.41} & 19.07 & 32.76 & 32.45 & \underline{3.81} & \underline{5.36} & 11.67 \\
 & iTransformer (2024) & 21.03 & 33.14  & 0.46 & 20.60 & 33.54 & 33.74 & 4.67 & 7.33 & 12.88 \\
 & STMFormer (2025) & 19.57 & 31.12 & 0.44 & 20.04 & 33.21 & 32.52 & 4.22 & 6.05 & 12.69 \\
 \midrule
\multirow{6}{*}{\rotatebox{90}{NODEs}} & STGODE (2021) & 21.40 & 33.47 & 0.45 & 20.53 & 33.46 & 33.83 & 4.56 & 6.71 & 13.01 \\
 & STG-NCDE (2022)& 21.21 & 33.80 & 0.45 & 21.33 & 35.64 & 34.53 & 4.37 & 6.62 & 13.12 \\
 & STDDE  (2024) & 22.85 & 34.26 & 0.46 & 21.04 & 34.36 & 34.15 & 4.44 & 6.57 & 12.96 \\
 & SGODE (2024) & 20.02 & 32.03 & 0.42 & 20.17 & 33.23 & 32.84 & 4.22 & 5.59 & 12.76 \\
 & AirPhyNet (2024) & 21.31$^*$ & 31.77$^*$ & 0.47$^*$ & 21.78 & 35.43 & 34.80 & 4.79 & 6.80 & 13.10 \\
 & AirDualODE (2025) & \underline{18.64}$^*$ & \underline{29.37}$^*$ & 0.42$^*$ & \underline{18.89} & \underline{32.26} & \underline{32.06} & 3.98 & 5.41 & 11.86 \\
\midrule
Ours & AirDDE & \textbf{16.92} & \textbf{27.78} & \textbf{0.38} & \textbf{17.03} & \textbf{29.91} & \textbf{30.82} & \textbf{3.53} & \textbf{4.87} & \textbf{10.94} \\ \bottomrule
\end{tabular}}
\caption{Results of AirDDE and baselines. The best results are bolded, and the second best results are underlined. The results with $*$ are cited from AirDualODE \cite{tian2024air}, while others are rerun using their official codes under multifactor settings.}
\label{tab:overall}
\end{table*}

\subsection{Spatiotemporal Decoder}
To consider spatial and temporal dependencies, we adopt the structure $\operatorname{GNN-GRU}(\cdot)$
(similar to the encoder) as the decoder. Given the adjacency matrix $\boldsymbol{A}$ learned from node embeddings (Eq.4) and the states $\boldsymbol{h}^{T+1:T+P}_{\text{p}}$ derived by the solver, the decoder is formulated as:
\begin{equation}
\hat{\boldsymbol{X}}^{t} = \operatorname{MLP}(\operatorname{GNN-GRU}(\boldsymbol{h}^{t}_{\text{p}},\boldsymbol{h}^{t-1}_{\text{p}},\boldsymbol{A}))
\end{equation}
where $t\in [T+1,T+P]$. $\hat{\boldsymbol{X}}^{T+1:T+P} \in \mathbb{R}^{N\times P}$ are the final predictions. 

For the training loss, we adopt the Huber loss, which is commonly used in ODE-based methods \cite{fang2021spatial,chen2024signed,long2024unveiling}.
Due to its robustness to outliers and smooth optimization behavior, the Huber loss is well-suited for modeling noisy systems, especially air quality dynamics, where measurement noise and extreme values are often significant. The loss is formulated as:
\begin{equation}
\mathcal{L}= 
\begin{cases}
\frac{1}{2}(\boldsymbol{X}-\hat{\boldsymbol{X}})^2 & ,|\boldsymbol{X}-\hat{\boldsymbol{X}}| \leq \delta \\ 
\delta|\boldsymbol{X}-\hat{\boldsymbol{X}}|-\frac{1}{2} \delta^2 & , \text { otherwise }
\end{cases}
\end{equation}
where $\delta$ is the threshold to change between delta-scaled L1 and L2 loss, which controls the sensitivity to outliers.

\section{Experiments}
\subsection{Experimental Setup}
\noindent \textbf{Datasets.}
We evaluate AirDDE on three real-world air quality datasets. KnowAir is provided by PM2.5GNN \cite{wang2020pm2}, including PM2.5 data and 17 meteorological factors from 184 cities across China. China-AQI and US-PM are provided by GAGNN \cite{chen2023group}. China-AQI includes AQI data and 7 meteorological factors from 203 cities across China. US-PM includes PM2.5 data and 7 meteorological factors from 175 counties across the US. The detailed statistics of these datasets are summarized in Table \ref{tab:dataset}. We follow the established preprocessing protocols from the original dataset studies. 

\noindent \textbf{Baselines.}
We compare AirDDE with 19 competitive baselines over 3 groups, including 
(1) \textbf{STGNNs}:
DCRNN \cite{li2017diffusion}, STGCN \cite{yu2018spatio}, ASTGCN \cite{guo2019attention}, MTGNN \cite{wu2020connecting}, PM25GNN \cite{wang2020pm2}, GAGNN \cite{chen2023group}, MegaCRN \cite{jiang2023spatio}, and HimNet \cite{dong2024heterogeneity};
(2) \textbf{Attentions}:
Crossformer \cite{zhang2023crossformer}, AirFormer \cite{liang2023airformer}, PDFormer \cite{jiang2023pdformer}, iTransformer \cite{liuitransformer}, and STMFormer \cite{li2025ssl};
and (3) \textbf{NODEs}: STGODE \cite{fang2021spatial}, STGNCDE \cite{choi2022graph}, SGODE \cite{chen2024signed}, STDDE \cite{long2024unveiling}, AirPhyNet \cite{hettige2024airphynet}, and AirDualODE \cite{tian2024air}.

\noindent \textbf{Settings.}
We split the datasets following the original dataset studies \cite{wang2020pm2,chen2023group}. 
KnowAir is divided chronologically for training, validation, and testing in a 2:1:1 ratio due to its ample four-year data span. China-AQI and US-PM are divided chronologically in a 7:1:2 ratio. 
All experiments are conducted on a single A100 GPU, employing the Adam optimizer with an initial learning rate of 0.005. We set the maximum number of epochs to 100 and employ an early stopping strategy with a tolerance of 10 epochs. For KnowAir, we set the batch size to 64, the input length to 24 (3-day), and the output length to 24 (3-day). For China-AQI and US-PM, we set the batch size to 32, the input length to 96 (4-day), and the output length to 24 (1-day). 
The number of global memory units is chosen from $\{8,16,32,64\}$. The time lag is chosen from $\{1,2,3,4\}$. We use the AutoML toolkit NNI \cite{nni} and its built-in Bayesian optimizer to efficiently tune the hyperparameters.

\subsection{Overall Comparison}
Table \ref{tab:overall} summarizes the results of all methods.
We can observe that:
(1) AirDDE achieves the best performance in all cases, outperforming the second-best with MAE reduction of 9.23\%, 9.85\%, and 7.3\%, on KnowAir, China-AQI, and US-PM, respectively. This improvement highlights the effectiveness of modeling continuous-time pollutant dynamics with delay effects.
(2) AirDualODE and PDFormer exhibit competitive results, as they address continuous-time dynamics and unified delay effects, respectively. AirDDE outperforms them by considering delays during continuous-time evolution, yielding more accurate forecasts with improved physical fidelity.
(3) AirDDE demonstrates the most significant improvement on China-AQI. Compared to KnowAir and US-PM, China-AQI features finer temporal granularity, higher pollution levels, and more locations, leading to the most complex pollutant dynamics. AirDDE effectively addresses this complexity, as it explicitly integrates transport paths based on distances and winds, achieving better adaptation to environments.

\subsection{Ablation Study}
\begin{table}[]
\centering
\resizebox{\linewidth}{!}{
\begin{tabular}{c|ccc|ccc|c}
\toprule
\multirow{2}{*}{Variant} & \multicolumn{3}{c|}{MAA} & \multicolumn{3}{c|}{PDE} & \multirow{2}{*}{AirDDE} \\
\cmidrule(lr){2-7}
 & w/o MAA & w/o GM & w/o LM & w/o PDE & w/o SST & w/o ATT &  \\
\midrule
1st & 15.45 & 15.13 & 15.37 & 15.97 & 15.70 & 15.00 & 14.53 \\
2nd & 20.18 & 18.47 & 17.80 & 19.93 & 18.86 & 19.82 & 17.16 \\
3rd & 21.65 & 19.57 & 18.54 & 21.47 & 19.79 & 21.33 & 18.01 \\
\midrule
AVG &  19.16 & 17.80 & 17.44 & 19.39 & 18.18 & 18.78 & 16.92 \\
\bottomrule
\end{tabular}}
\caption{MAE results of the ablation study on KnowAir.}
\label{tab:ablation}
\end{table}

\noindent \textbf{MAA Module.} 
We design three variants: (1) Removing the entire module (-w/o MAA) and using an MLP to encode inputs to get the initial state \cite{long2024unveiling}. (2) Removing the global memory modeling (-w/o GM). (3) Removing the local memory modeling (-w/o LM).
As shown in Table \ref{tab:ablation}, AirDDE outperforms its variants -w/o MAA, -w/o GM, and -w/o LM, showing the contributions of memory augmentation and both global and local historical patterns to delay modeling.
In addition, -w/o MAA causes a significant drop in 3rd-day forecasts, where historical pollution patterns are more influential, highlighting MAA's crucial role in long-term forecasting.
 
\noindent \textbf{PDE Function.}
We design three variants: (1) Removing the entire module (-w/o PDE) and directly feeding encoder states to the decoder. (2) Removing the source/sink term (-w/o SST). (3) Removing the physics priors and replacing with attention-based evolving function \cite{tian2024air} (-ATT).
As shown in Table \ref{tab:ablation}, AirDDE performs better than -w/o PDE, -w/o SST, and -ATT, showing the effectiveness of the delay-aware evolution, multifactor enhancement, and physics priors, respectively, for continuous-time modeling.
In addition, the physics-guided variants, i.e., -SST and AirDDE, outperform purely data-driven variants, i.e., -w/o PDE and -ATT.
This is because the physical priors offer structural guidance to capture delay effects and maintain consistency with real-world pollutant dynamics.

\subsection{Long-Term Study}

\begin{figure}
    \centering
    \includegraphics[width=0.7 \linewidth]{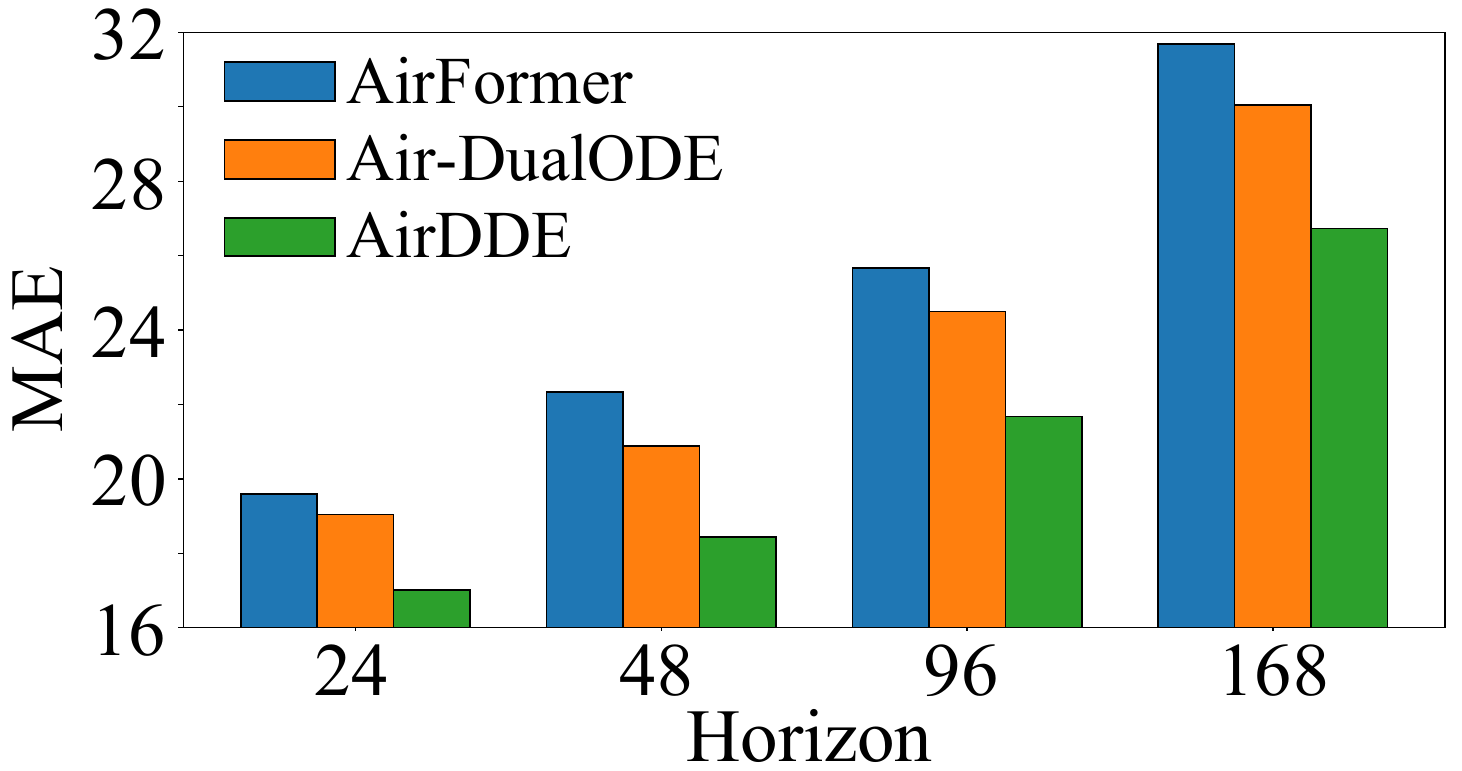}
    \caption{Results of the long-term study on China-AQI.}
    \label{fig:long-term}
\end{figure}

To evaluate the long-term forecasting ability of AirDDE, we fix the input length to $T=96$ and extend the output horizon to $H = \{24, 48, 96, 168\}$. As shown in Fig. \ref{fig:long-term}, AirDDE consistently achieves the best performance, with its advantage becoming more pronounced at longer forecasting horizons. 
These results demonstrate its superior long-term forecasting capability, highlighting the effectiveness of delay-aware continuous modeling with physical priors over extended horizons.

\begin{table}[]
\centering
\resizebox{\linewidth}{!}{
\begin{tabular}{c|cccc|cccc}
\toprule
\multirow{2}{*}{Dataset} & \multicolumn{4}{c|}{Time lag} & \multicolumn{4}{c}{Number of global   memory unit} \\
\cmidrule(lr){2-9}
 & 0 & 1 & 2 & 3 & 8 & 16 & 32 & 64 \\
\midrule
China-AQI & 18.54 & 17.03 & 19.87 & 21.05 & 19.66 & 18.49 & 17.03 & 18.78 \\
US-PM & 4.32 & 3.98 & 3.53 & 4.06 & 3.93 & 3.53 & 4.07 & 4.25 \\
\bottomrule
\end{tabular}}
\caption{Results of the hyperparameter study on China-AQI.}
\label{tab:hyper}
\end{table}

\begin{figure}
    \centering
    \includegraphics[width=\linewidth]{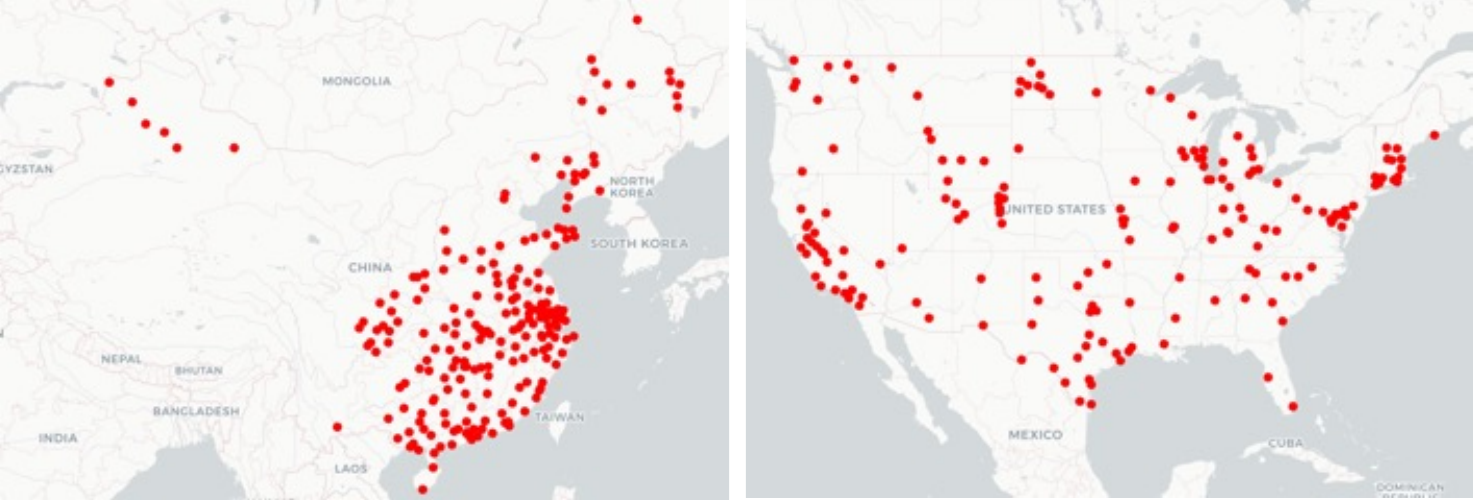}
    \caption{Station distribution of China-AQI (left) and US-PM (right).}
    \label{fig:location}
\end{figure}

\subsection{Hyperparameter Study}

We evaluate the effect of key hyperparameters in AirDDE. To ensure fairness, all other hyperparameters are held fixed when varying a specific one.

\noindent \textbf{Time Lag $\tau$.} 
$\tau$ governs the construction of transport paths and directly influences the advection process. We vary $\tau$ from 0 to 3 with a step size of 1.
As summarized in Table \ref{tab:hyper}, the optimal values are $\tau=1$ on China-AQI and $\tau=2$ on US-PM. This is because a too-small $\tau$ fails to capture sufficient transport delays, whereas an overly large $\tau$ introduces outdated or less relevant historical states.
In addition, the discrepancy of $\tau$ between the two datasets reflects dataset-specific pollutant dynamics.
As illustrated in Fig.~\ref{fig:location}, the denser station distribution in China-AQI results in rapid and localized pollutant propagation, favoring shorter time lags. In contrast, the sparser station distribution in US-PM requires longer $\tau$ to model long-range transport delays.

\noindent \textbf{Number of Global Memory Units $m$.} 
$m$ controls the model's capacity to store global historical patterns.
We evaluate $m$ in $\{8,16,32,64\}$. As summarized in Table \ref{tab:hyper}, the optimal values are $m=32$ for China-AQI and $m=16$ for US-PM, while both larger and smaller values degrade performance.
This is because a too-small $m$ fails to capture diverse patterns, whereas an overly large $m$ introduces redundancy or noise, increasing the risk of overfitting.
In addition, the discrepancy of $m$ reflects the different characteristics of the two countries. 
Due to higher industrial activity and denser population centers, air quality patterns in China exhibit more complex pollution dynamics than those in the US, requiring greater memory capacity.

\subsection{Efficiency Study}

\begin{table}[]
\centering
\resizebox{0.8 \linewidth}{!}{
\begin{tabular}{c|c|ccc}
\toprule
Method & Conf. & \begin{tabular}[c]{@{}c@{}}GPU Memory\\ (GB) \end{tabular} & \begin{tabular}[c]{@{}c@{}}Training Time\\ (Min/Epoch)\end{tabular} & MAE \\
\midrule
STGODE & KDD 2021 & 14.88 & 6.37 & 20.53 \\
STG-NCDE & AAAI 2022 & 5.89 & 39.14 & 21.33 \\
AirFormer & AAAI 2023 & 8.78 & 2.95 & 19.60 \\
PDFormer & AAAI 2023 & 15.71 & 5.04 & 19.07 \\
STDDE & WWW 2024 & 21.43 & 24.64 & 21.04 \\
SGODE & AAAI 2024 & 12.22 & 11.06 & 20.17 \\
AirPhyNet & ICLR 2024 & 14.31 & 4.78 & 21.78 \\
AirDualODE & ICLR 2025 & 11.14 & 10.09 & 18.89 \\
\midrule
AirDDE & Ours & 10.46 & 9.24 & 17.03 \\
\bottomrule
\end{tabular}}
\caption{Results of the efficiency study on China-AQI.}
\label{tab:effciency}
\end{table}

We compare the GPU memory usage, training time, and MAE of AirDDE against competitive baselines and NODEs on the China-AQI dataset. As shown in Table \ref{tab:effciency}, we can observe that:
(1) AirDDE achieves the best prediction performance with relatively reasonable computational cost. Although the second-best method, AirDualODE, also shows strong accuracy, its dual ODE solvers introduce heavy overhead and reduce efficiency.
(2) Compared with conditioned NODEs, i.e., STG-NCDE and STDDE, AirDDE improves in efficiency and accuracy. This is mainly because AirDDE incorporates transport paths directly into the advection term, avoiding the expensive continuous-path encoding while still modeling delay-aware dependencies.
(3) Compared with competitive baselines, i.e., AirFormer and PDFormer, AirDDE incurs longer training time due to historical state maintenance for delay modeling. However, it achieves significant MAE reductions, i.e., 13.11\% over AirFormer and 10.70\% over PDFormer. Moreover, AirDDE remains more memory-efficient than PDFormer, demonstrating a favorable efficiency–performance trade-off.

\subsection{Robustness Study}

\begin{table}[]
\centering
\resizebox{\linewidth}{!}{
\begin{tabular}{c|c|ccc|ccc}
\toprule
\multirow{2}{*}{Method} & \multirow{2}{*}{Original} & \multicolumn{3}{c|}{Missing Rate} & \multicolumn{3}{c}{SNR}  \\
\cmidrule(lr){3-8}
& & 10\% & 30\% & 50\% & 80db & 60db & 40db \\
\midrule
AirFormer & 19.17 & 23.09 & 27.83 & 38.44 & 24.68 & 31.85 & 45.84  \\
PDFormer & 19.06 & 22.43 & 29.86 & 41.23 & 25.86 & 36.93 & 48.42\\
STDDE & 22.85  & 26.11 & 32.54 & 40.78 & 28.75 & 35.92 & 49.34 \\
AirPhyNet & 21.31 & 25.20 & 31.30 & 37.79 & 27.82 & 33.60 & 46.42  \\
AirDualODE & 18.64 & 21.10 & 25.83 & 34.82 & 23.46 & 31.14 & 38.42 \\
\midrule 
AirDDE & 16.92 & 18.39 & 22.44 & 29.45 & 21.02 & 26.68 & 32.51 \\
Impro. & 9.23\% & 12.84\% & 13.12\% & 15.42\% & 14.06\% & 14.32\% & 15.38\% \\
\bottomrule
\end{tabular}}
\caption{MAE Results of the robustness study on KnowAir.}
\label{tab:robust}
\end{table}

In real-world scenarios, air quality data are often irregular due to sensor failures and noise. To evaluate the robustness of AirDDE, we compare it with competitive baselines on KnowAir.
For missing data, following STNCDE \cite{choi2022graph}, we randomly drop 10\% to 50\% of values for each sensor independently. For noisy data, following CrossGNN \cite{huang2023crossgnn}, we inject Gaussian white noise with varying intensities, progressively decreasing the signal-to-noise ratio (SNR) from 80 dB to 40 dB. 
As shown in Table \ref{tab:robust},  AirDDE consistently outperforms all baselines under both settings, with MAE improvements increasing as data quality deteriorates, i.e., from 9.23\% to 15.42\% with higher missing rates, and from 14.06\% to 15.38\% under stronger noise.
This trend highlights AirDDE’s superior robustness, particularly under more challenging conditions. This advance stems from its global memory and physics-guided evolution, which enables effective recovery and denoising via global historical patterns while maintaining physical consistency with real-world pollutant dynamics.



\subsection{Case Study}
\begin{figure}
    \centering
    \includegraphics[width=\linewidth]{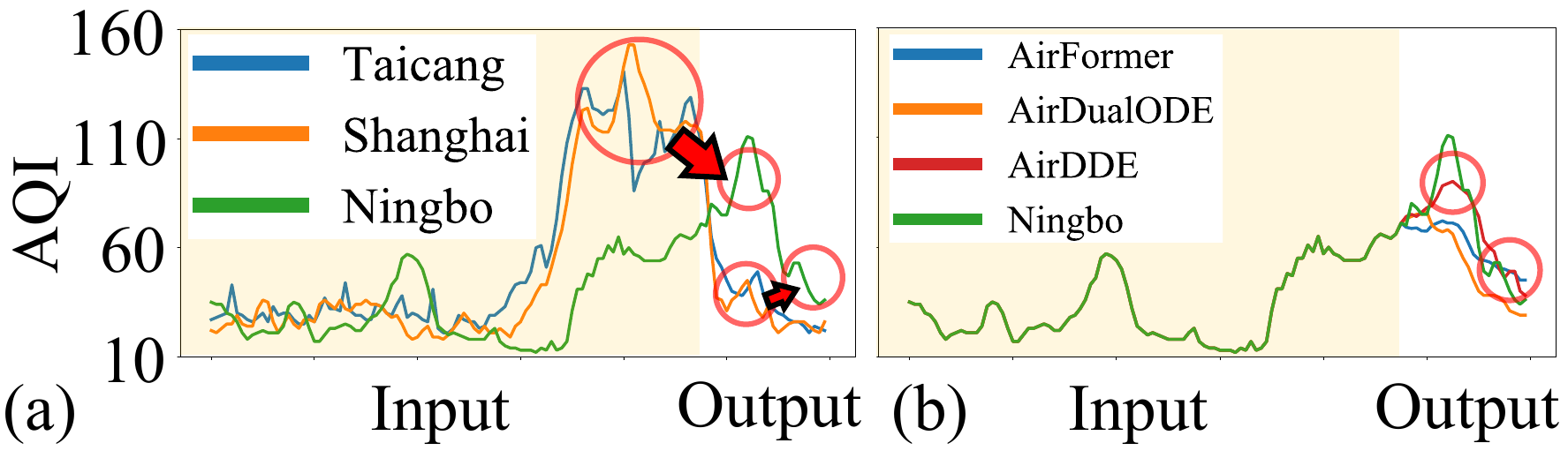}
    \caption{Case of city-wise advection with delay effects.}
    \label{fig:case1}
\end{figure}

\noindent \textbf{City-Wise Advection with Delay Effects.}
Fig. \ref{fig:case1} (a) illustrates the AQI of Taicang, Shanghai, and Ningbo from Nov.14 10:00, 2018 to Nov.19 10:00, 2018 in the test dataset of China-AQI. As shown in red cycles, since the average wind direction in Shanghai and Taicang during Nov.15 and Nov. 18 is north and northeast, the pollutant is driven to the downstream city, i.e, Ningbo, resulting in AQI peaks with lags in Ningbo.
As shown in the predicted results in Fig. \ref{fig:case1} (b), compared to SOTA methods, AirDDE can effectively capture these peaks with lags, as AirDDE explicitly considers this wind-driven delays in the advection process.

\begin{figure}
    \centering
    \includegraphics[width=0.9 \linewidth]{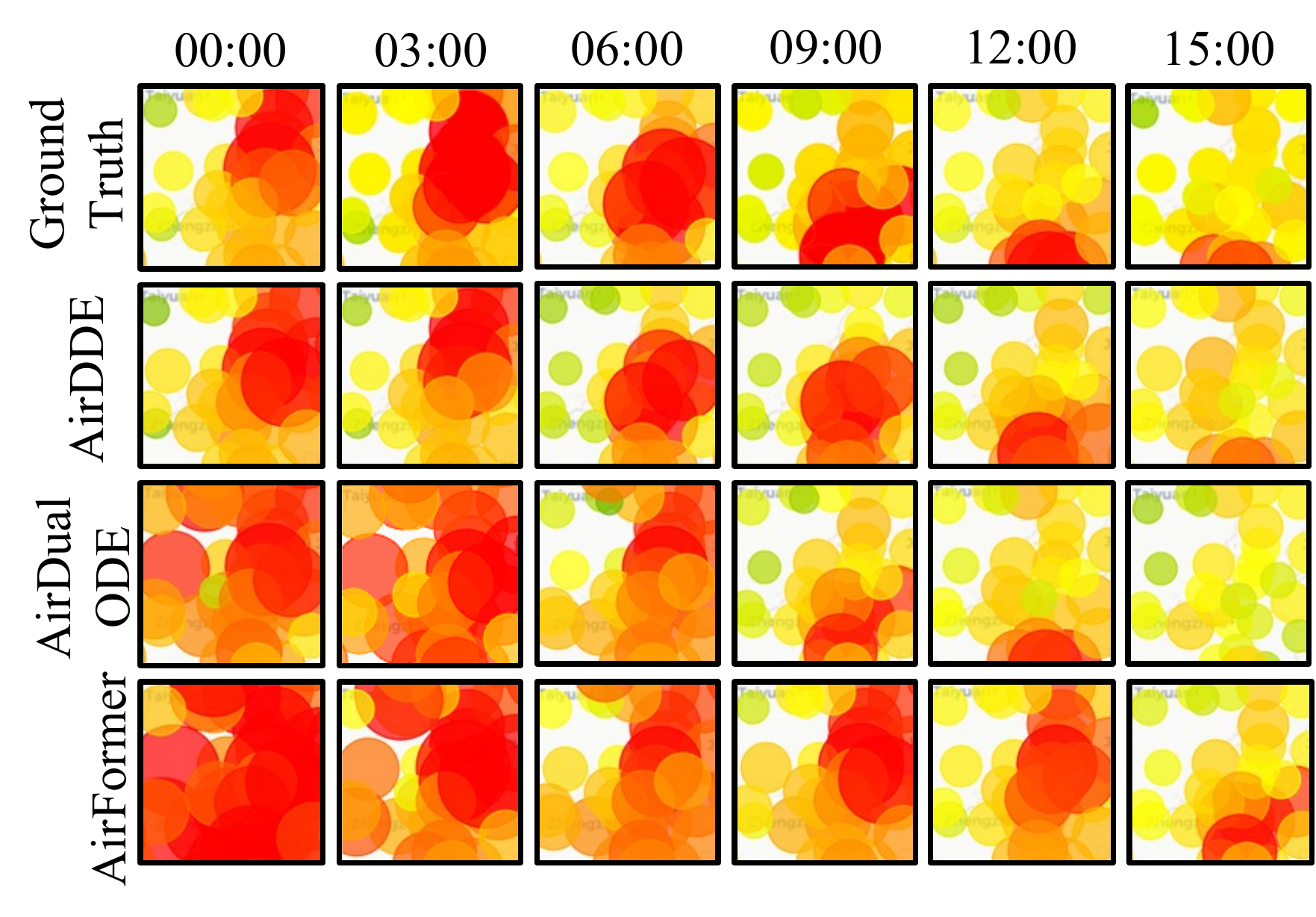}
    \caption{Case of region-wise advection with delay effects.}
    \label{fig:case2}
\end{figure}

\noindent \textbf{Region-Wise Advection with Delay Effects.} 
Figure \ref{fig:case2} illustrates the PM2.5 concentration of a region in Shanxi Province from 00:00 to 15:00 on December 3, 2018, using data from the KnowAir test dataset.
Each circle represents the PM2.5 concentration at a location, with a larger radius and deeper red color indicating higher concentration levels.
During this period, wind-driven advection transports pollutants from northeast to southwest. Downwind areas receive pollutants from upwind regions after a certain time lag, reflecting the inherent delay effect. Compared to predicted results from SOTA methods, AirDDE effectively captures the transport path of pollutants, showcasing its ability to represent delay-aware pollutant dynamics.

\section{Conclusions and Future Work}
In this work, we introduce AirDDE, the first neural delay differential equation framework that captures delay effects during continuous-time pollutant evolution. We introduce two novel blocks guided by physics priors: the MAA module to capture delay effects modulated by historical multifactor data and the PDE function to capture delay-aware pollutant accumulation patterns. 
Extensive experimental results demonstrate that AirDDE outperforms 19 competitive baselines, reducing average MAE by 8.79\% over the best baselines, while demonstrating its practical strength in long-term forecasting and robustness.

Despite such advantages, several open directions remain for more comprehensive delay modeling. We highlight three key areas:
(1) improving the efficiency of delay state maintenance;
(2) incorporating uncertainty into delay estimation, given the stochastic nature of wind fields; and
(3) modeling compound delays to capture complex transport paths involving intermediate regions.


\clearpage
\bibliography{ref}

\end{document}